\documentclass[runningheads]{llncs}
\usepackage[T1]{fontenc}
\usepackage{graphicx}
\usepackage{cite}
\usepackage{amsmath,amssymb,amsfonts}
\usepackage{algorithmic}
\usepackage{graphicx}
\usepackage{textcomp}
\usepackage{graphicx} 
\usepackage{epsfig} 
\usepackage{amsmath} 
\usepackage{amssymb}  
\usepackage{booktabs}        
\usepackage{graphicx}        
\usepackage{multirow}        
\usepackage{pifont}          
\usepackage{array}           
\usepackage{makecell}        
\usepackage{cite}
\usepackage{pifont}  
\usepackage{booktabs}  
\usepackage{colortbl}
\usepackage[table]{xcolor}
\definecolor{Gray}{gray}{0.65}
\usepackage{bm}
\usepackage{microtype}
\usepackage{subcaption}

\begin{document}
\title{ClipTBP: Clip-Pair based Temporal Boundary Prediction with Boundary-Aware Learning for Moment Retrieval}
\titlerunning{ClipTBP}
%
\author{Ji-Hyeon Kim\orcidID{0009-0005-7319-4664} \and
Ho-Joong Kim\orcidID{0000-0003-4200-5136} \and
Seong-Whan Lee\orcidID{0000-0002-6249-4996}}
\authorrunning{J-H Kim et al.}
%
\institute{
Dept. of Artificial Intelligence, Korea University, Seoul 02841, Republic of Korea \\
\email{\{j\_h\_kim, hojoong\_kim, sw.lee\}@korea.ac.kr}
}
\maketitle              
\begin{abstract}
Video moment retrieval is the task of retrieving specific segments of a video corresponding to a given text query. Recent studies have been conducted to improve multimodal alignment performance through visual-linguistic similarity learning at the snippet-level and transformer-based temporal boundary regression. However, existing models do not calculate similarity by considering the relationships between multiple answer segments that match the query. Therefore, existing models are easily influenced by visually similar segments in the surrounding context. Existing models calculate similarity at the snippet-level and ignore the relationships between multiple answer segments corresponding to a single query. Therefore, they struggle to exclude segments irrelevant to the query. 
To address this issues, we propose ClipTBP, a clip-pair temporal boundary prediction framework based on boundary-aware learning. ClipTBP introduces a clip-level alignment loss for explicitly learning the semantic relationship between answer segments.  ClipTBP also predicts accurate temporal boundaries by applying both main boundary loss and auxiliary boundary loss. ClipTBP consistently improves performance when applied to various existing models and demonstrates more robust boundary prediction performance even in ambiguous query scenarios.

\keywords{ Boundary-aware learning \and Clip-level similarity \and Contrastive learning \and Representation learning \and Video moment retrieval.}
\end{abstract}
\section{Introduction}
Computer vision \cite{mr_1,mr_2,mr_4,luo2024zero,kim2021fre,jeong2020multimodal,Lee_2020_CVPR,MIN2022439} in recent years has expanded beyond image recognition to include video understanding. Videos have temporal continuity, unlike static images, and therefore address complex characteristics that require understanding of both scene context and event progression. 
Video moment retrieval \cite{mr_1,mr_2,mr_3,cal,cv_1} is a task of retrieving specific video segments corresponding to a given text query, and it is an essential method in various fields that automatically analyze and utilize videos. Video moment retrieval is used in video search engines, content recommendation systems, and video summarization. One of the main challenges of video moment retrieval is accurately predicting the start and end boundaries of the search target segment. Another important challenge is understanding the semantic connection between text queries and videos and taking into account the temporal relationship within the video. The two model structures mainly used are the DETR-based \cite{detr} models, which are end-to-end transformer-based detection frameworks \cite{qd-detr, tr-detr, td-detr,cg-detr}, and snippet-level offset regression models \cite{lin2023univtg, video-mamba-suite, cao2024flashvtg}.

\begin{figure}[t]
    \centering
    \includegraphics[width=0.82\linewidth]{figures/figure1}
    \vspace{-6pt}
    \caption{\textbf{Challenges when multiple answer segments exist in a video.} The blue boxes are the answer segments matching the query in the ground truth. The red boxes at the top are the segments predicted by baseline model (FlashVTG \cite{cao2024flashvtg}) and the red boxes at the bottom are the segments predicted by our proposed model. The images connected to the green arrows are the answer segments that match the query, and the images connected to the orange arrows are segments unrelated to the query.} 
    \label{intro fig}
    \vspace{-10pt}
\end{figure}

Existing moment retrieval models \cite{lin2023univtg, cao2024flashvtg, video-mamba-suite, td-detr} directly calculate the similarity between video snippets divided into fixed-length segments and text queries, then predict the temporal offsets of candidate segments based on this similarity. This learning structure treats each snippet as an independent unit, eliminating the need to encode the entire video as a single long sequence. Thus, these models achieve high computational efficiency and stable learning and inference even for long videos. The explicit query-video comparison performed at the snippet-level enables intuitive alignment. However, this snippet-independent learning structure has limitations in reflecting the semantic relationships and global temporal context among multiple correct answer segments for a single query. As a result, when segments appear that are visually similar to the answer segment but irrelevant to the query, the model is unable to distinguish them.

To illustrate this issue, we compare the moment retrieval results of our model with the baseline model (FlashVTG \cite{cao2024flashvtg}) when there are multiple answer segments in the ground truth (GT). Fig.~\ref{intro fig} (i) shows a woman with pink hair, matching the query \textit{A woman with pink hair}. However, this scene does not depict the query \textit{describing her trip}, but instead shows a woman walking around during her trip. The baseline model incorrectly predicts the segment as the answer segment due to visual similarity, despite the scene not fully matching the query, but only partially matching some content of the query. Furthermore, Fig.~\ref{intro fig} (ii) and (iii) show the failure cases of the baseline model in precise boundary regression. Although the segments in (ii) and (iii) are visually dissimilar to the ground-truth answer segments, the baseline model fails to recognize a brief temporal gap between adjacent answer segments, resulting in overly broad boundary predictions. This limitation stems from the lack of fine-grained representation learning, which results in a single prediction that spans multiple answer segments.


In this paper, we propose a clip-pair based temporal boundary prediction with boundary-aware learning for moment retrieval (ClipTBP). ClipTBP uses three additional losses to strengthen temporal boundary prediction. Unlike existing methods that depend on snippet-query matching, our model additionally learns the relationships between answer segments. 
ClipTBP learns the semantic relationships and consistency between answer segments in videos, allowing for enhanced moment retrieval by improving the accuracy of temporal boundaries of answer segments. Our proposed model is particularly efficient when events matching queries occur repeatedly. In this case, modeling temporal and semantic dependencies is required to predict the boundaries of each answer segment.
While existing models focus on the similarity between each snippet and query and have difficulty distinguishing adjacent answer segments and segments that are visually similar to answer segments, ClipTBP addresses this issue by explicitly learning the semantic relationships between answer segments. In addition, through boundary-aware learning, it encourages precise recognition of segment boundaries in the learning process, allowing precise temporal boundary prediction for adjacent segments. On QVHighlights \cite{moment-detr} and TACoS \cite{tacos}, ClipTBP shows consistent performance improvement when applied to various snippet-level offset regression-based models and even to DETR-based models. These results indicate that our ClipTBP is easily integrated into various existing moment retrieval methods.

In summary, this paper includes the following three contributions:
\begin{itemize}
    \item We propose ClipTBP that leads to improved moment retrieval performance through learning the relationship between multiple answer segments and  learning the boundary of the answer segments.
    \item Our proposed method enhances boundary learning through advanced query-snippet alignment using inter-segments similarity and precise learning inside and outside of the boundary.
    \item Our proposed method seamlessly integrates into various moment retrieval models including both snippet-level and DETR-based frameworks, consistently improving performance on QVHighlights and TACoS.
\end{itemize}
\section{Related Works}


\noindent \textbf{Video moment retrieval} aims to localize specific time segments in an untrimmed video corresponding to a given text query. This task involves accurately modeling the semantic relationship between the text query and the video segment, and accurately predicting the start and end boundaries of the relevant video moment.
Most transformer-based approaches have been studied in this field. Moment-DETR \cite{moment-detr} removes the dependency on explicit proposals by organizing moment retrieval as an end-to-end transformer-based set prediction task. UMT \cite{liu2022umt} proposes an integrated multimodal transformer architecture that effectively integrates multimodal context by jointly performing moment retrieval and highlight detection. 
More recently, FlashVTG \cite{cao2024flashvtg} improves performance by modeling multi-scale temporal context and improving prediction through adaptive scoring. Video Mamba Suite \cite{video-mamba-suite} improves computational efficiency over transformer models by adding blocks \cite{vim,ssm} to UniVTG \cite{lin2023univtg}. TD-DETR \cite{td-detr} improves prediction performance by introducing a temporal decoder conditioned on text queries, enabling effective inference of query-relevant temporal information throughout the video.
However, the existing method mainly focuses on the similarity between queries and each individual clip, so it has difficulty in searching for situations where visually similar events occur repeatedly within a video. In contrast to snippet-level similarity modeling, we propose a method that enables accurate temporal boundary prediction in challenging scenarios by additionally modeling the similarity between answer segments.


\noindent \textbf{Similarity-based learning} is a methodology that measures the similarity between data to effectively learn the embedding space and, based on this, perform accurate predictions and retrieval. Representative similarity-based loss functions such as contrastive loss \cite{contrastive_learning} and triplet loss \cite{triplet_loss} are mainly used to construct feature vectors that clearly distinguish between positive and negative relationships between data samples. Recent studies \cite{panta2024cross,pan2025fawl,zhang2021video} suggest that this method effectively strengthens cross-modal alignment in video moment retrieval.
In particular, learning the similarity relationship at the snippet-level helps to capture the semantic relationship between video clips and text queries even in cases where there are ambiguous queries or repeated events. Through explicit supervision of semantic alignment between answer segments, it improves the ability to distinguish between multiple candidate segments with similar visual representations or meanings. In this paper, we extend the advantages of this similarity-based learning by combining clip-level alignment learning and boundary-aware learning to effectively integrate semantic matching and precise temporal segment prediction.



\begin{figure*}[!t]
    \centering
    \includegraphics[width=\linewidth]{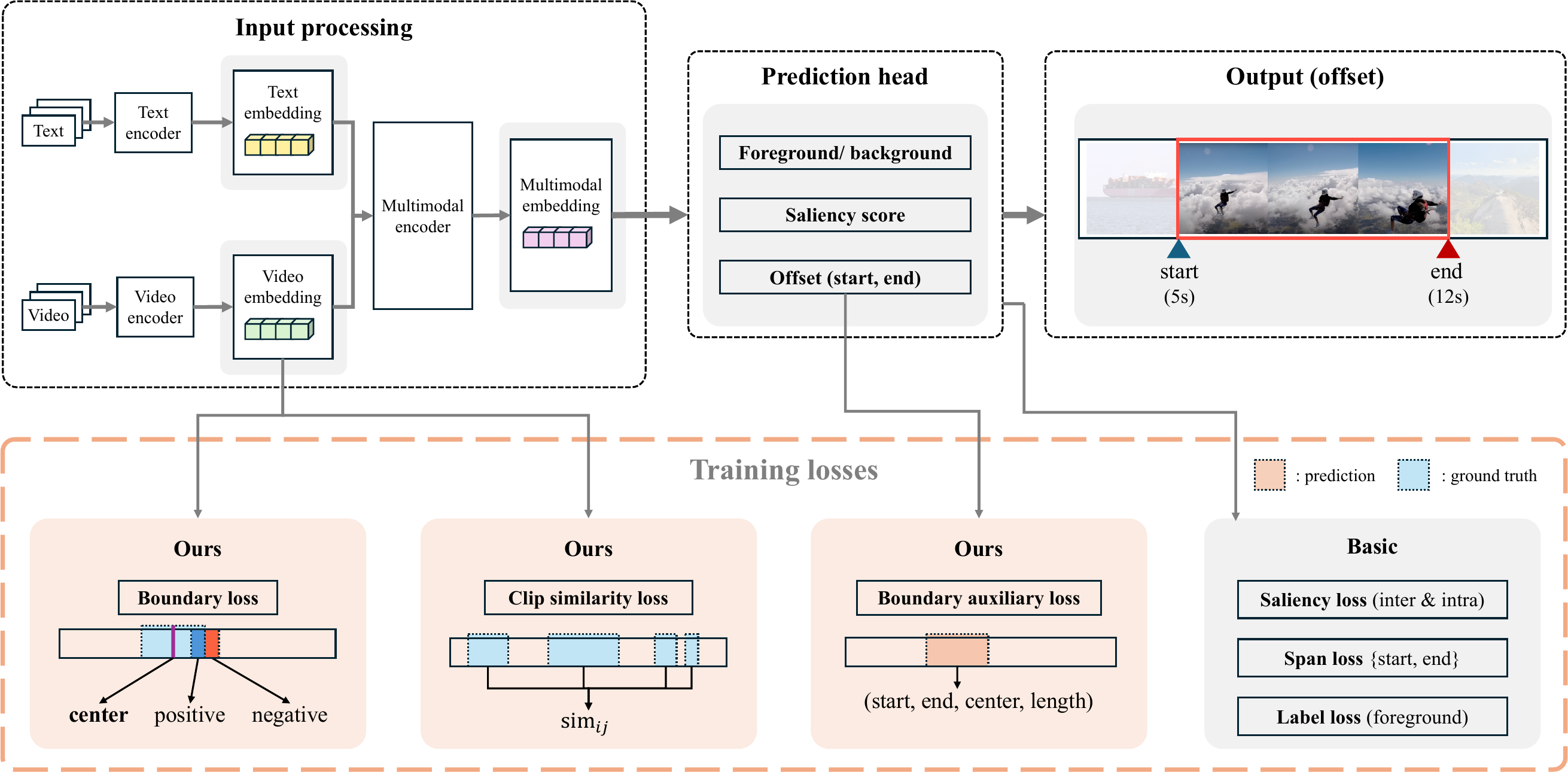}
    \vspace{-15pt}
    \caption{\textbf{Overall framework of {ClipTBP}.} ClipTBP encodes the input video $V=\{v_t\}^T_{t=1}$ and text query $q$ into a video encoder \( f_v(\cdot) \) and text encoder \( f_q(\cdot) \), respectively, and then passes them through a multimodal encoder \( f_m(\cdot) \) to generate snippet-level multimodal representations \( \mathbf{Z} = \{ \mathbf{z}_i \}_{i=1}^{T} \). The prediction head generates a saliency score, start/end offset, and foreground probability for each clip. The training uses the basic losses of the existing models, along with clip-level similarity loss for aligning the meanings of answer segments, boundary loss for distinguishing the boundaries of segments, and auxiliary boundary loss for stabilizing regression.}
    \vspace{-10pt}
    \label{method fig}
\end{figure*}
\section{Method}
In this section, we describe the proposed method, ClipTBP. ClipTBP is a framework that explicitly models the relationship between answer segments in GT and improves temporal boundary prediction performance through additional learning on the boundaries of answer segments. Unlike existing methods that mainly focus on the similarity between queries and snippets, we use clip-level similarity loss to learn the semantic structure between answer segments. We also directly learn the boundaries of answer segments and their offsets to improve boundary prediction. The entire framework of ClipTBP is illustrated in Fig.~\ref{method fig}.

\subsection{Preliminaries}
Video moment retrieval is a task that aims to localize a temporal segment in a video that semantically corresponds to a natural language query. We denote the video set as $\mathcal{X}=\{x_n\}_{n=1}^{N_v}$, where $N_v$ is the number of videos, and the query set as $Q=\{q_m\}_{m=1}^{N_q}$, where $N_q$ is the number of queries. 
The objective of the model is to predict the GT timestamps $\mathcal{T}=\{t_s,t_e\}$ matching the given query, where $t_s$ is the GT start timestamp and $t_e$ is the GT end timestamp.
The input video $\mathcal{X}$ and query $\mathcal{Q}$ are passed through a video encoder $f_v(\cdot)$ (e.g., SlowFast \cite{feichtenhofer2019slowfast}) and a text encoder $f_q(\cdot)$ (e.g., CLIP \cite{clip}) to obtain embeddings $\mathbf{V}=\{\mathbf{v}_i\}^T_{i=1}$ and $\mathbf{Q}=\{\mathbf{q}_i\}^{L_q}_{i=1}$, where $T$ is the length of the video, $L_q$ is the number of tokens in the query and $i$ is the index. $\mathbf{v}_i\in\mathbb{R}^D$ and $\mathbf{q}_i\in\mathbb{R}^D$, where $D$ is the shared embedding dimension.
These embeddings are then combined with type and positional embeddings and input into the multimodal encoder $f_m(\cdot)$, generating multimodal embeddings $\mathbf{Z}=\{\mathbf{z}_i\}^T_{i=1}$ that reflect the spatio-temporal interactions between text and video, where $\mathbf{z}_i\in\mathbb{R}^D$.

\subsection{Clip-Pair based Temporal Boundary Prediction}
ClipTBP uses three main loss functions in addition to the loss functions of the existing snippet-level offset regression models to learn semantic alignment between clips and precise boundary prediction at the same time: clip-level similarity loss, main boundary loss, and auxiliary boundary loss. Each loss provides a different learning signal, and all losses are combined into a final total loss to optimize.

\noindent \textbf{Clip-level similarity loss $\mathcal{L}_{clip}$}
The clip-level similarity loss is used to learn the semantic relationship between clips in contrastive learning, and through this, ClipTBP directly learns the relationship between answer segments structure throughout the video. In particular, clip pairs that exist within the same GT segment are considered positive, and other pairs are considered negative. Among the negatives, only the top-$k$ with the highest similarity are selected for hard negative mining. The cosine similarity ${s}_{i,j}$ between two clips $\mathbf{v}_i$ and $\mathbf{v}_j$ is calculated as follows:
\begin{equation}
 s_{i,j}=\frac{\mathbf{v}_i^\top \mathbf{v}_j}{\|\mathbf{v}_i\|\,\|\mathbf{v}_j\|}\text{,}
\end{equation}
where $\mathbf{v}_i$ is the embedding of the clip at index $i$.
Given the ground truth normalized timestamps $(t_s,t_e)$ and the clip-index set $\mathcal{I}=\{1,\dots,T\}$, 
we define the positive pair set as $\mathcal{P}=\{(i,j)\mid i,j\in\mathcal{I},\ \ t_s \le i/T \le t_e,\ \ t_s \le j/T \le t_e\}$, and the negative pair set as $\mathcal{N}=\{(i,j)\notin\mathcal{P\}}$. 
The positive pairs correspond to clips that share similar semantics within the GT answer segments matching the given query, while negative pairs correspond to clips outside the segment that do not share such semantics.
For hard negative mining, we further select the top-$k$ pairs with the largest similarity values from $\mathcal{N}$, denoted as $\mathcal{N}_k$.
The final clip similarity loss is then defined as follows: 
\begin{equation}
\mathcal{L}_{clip} = -\log\left(
\frac{
\sum\limits_{(i,j)\in\mathcal{P}} \exp\!\left( s_{i,j}/\tau \right)
}{
\sum\limits_{(i,j)\in\mathcal{P}} \exp\!\left( s_{i,j}/\tau \right)
+\sum\limits_{(i,j)\in\mathcal{N}_k} \exp\!\left( s_{i,j}/\tau \right)
}
\right),
\end{equation}
where $\tau$ is the temperature parameter.

\noindent \textbf{Main boundary loss $\mathcal{L_\text{boun}}$}
The main boundary loss is used to learn the distinction between the boundary just inside and outside the GT segment, guiding the clip to predict the boundary information of the accurate answer segment. In detail, clips around the beginning, center, and end of the GT segment are set as positive, and clips located outside the segment are set as negative, and contrastive loss based on the difference in similarity between clips is configured. It is formulated as a margin-based ranking loss, using a dynamically set margin depending on the length of the GT segment. This enables precise boundary prediction even when ambiguous or similar scenes are clustered around. The cosine similarity between each clip $\mathbf{v}_i$ and the center of the GT answer clip $\mathbf{v}_c$ is $s_{c,i}$, where $\mathbf{v}_c$ is center of each GT answer segment. Given the positive and negative index sets $\mathcal{I}_{\text{pos}}$ and $\mathcal{I}_{\text{neg}}$, 
the average similarities with respect to the center clip are defined as follows:
\begin{equation}
    s_\text{pos}=\frac{1}{|\mathcal I_{\text{pos}}|}\sum_{i\in\mathcal{I}_\text{pos}} s_{c,i},\quad
    s_\text{neg}=\frac{1}{|\mathcal I_{\text{neg}}|}\sum_{i\in\mathcal{I}_\text{neg}} s_{c,i}\text{.}
\end{equation}
Using the dynamically adjusted margin $\delta$, the boundary loss is defined as follows:
\begin{equation}
    \mathcal{L}_{boun}=\max\!\big(0,\ s_\text{neg}-s_\text{pos}+\delta\big)\text{,}
\end{equation}
where $\delta=\min\!\big(0.3,\ 0.1+0.1\cdot\tfrac{j_e-j_s}{T}\big)$ is dynamically adjusted according to the segment length normalized by the number of clips $(j_e-j_s)/T$. The $j_s$ and $j_e$ denote the start and end indices of the GT answer segment for the query.

\noindent \textbf{Auxiliary boundary loss $\mathcal{L}_{\text{b\_aux}}$}
The auxiliary boundary loss is a supplementary loss designed to help the model perform boundary regression stably during the training process. While the main boundary loss contrastively learns the similarity structure between clips near the ground truth boundaries, this auxiliary loss provides precise regression supervision for the center position and interval length in addition to the predicted start and end timestamps, thereby strengthening the structural consistency of the boundary representation. By providing additional supervision signals for relatively stable features such as center or length, the overall regression learning is smoothly guided. 
Given the predicted start and end timestamps $(\hat{t}_s,\hat{t}_e)$ and the GT answer timestamps $(t_s,t_e)$, the center positions and segment lengths are defined as follows:
\begin{equation}
    \hat{t}_c=\frac{\hat{t}_s+\hat{t}_e}{2}\text{,} \quad t_c=\frac{t_s+t_e}{2}\text{,}
\end{equation}
\begin{equation}
    \hat{l}=\hat{t}_e-\hat{t}_s\text{,} \quad l=t_e-t_s\text{.}
\end{equation}
The regression targets consist of four elements: start $(\hat{t}_s,t_s)$, end $(\hat{t}_e,t_e)$, 
center $(\hat{t}_c,t_c)$, and length $(\hat{l},l)$.
Each of these elements provides complementary information about the boundary structure, 
and they are optimized using the SmoothL1 \cite{smooth_l1} with different weights. 
The final auxiliary boundary loss is defined as follows:
\begin{equation}
\begin{split}
    \mathcal{L}_{b\_aux} &=
    \lambda_s\cdot\text{SmoothL1}(\hat{t}_s,t_s)
    +\lambda_e\cdot\text{SmoothL1}(\hat{t}_e,t_e) \\
    &\quad+\lambda_c\cdot\text{SmoothL1}(\hat{t}_c,t_c)
    +\lambda_l\cdot\text{SmoothL1}(\hat{l},l)\text{.}
\end{split}
\end{equation}

\subsection{Training and Inference}

ClipTBP improves both alignment and boundary prediction performance during training by combining the basic loss functions of existing snippet-level offset regression models with the three loss functions proposed in this paper: clip-level similarity loss $\mathcal{L}_{clip}$, main boundary loss $\mathcal{L}_{boun}$, and auxiliary boundary loss $\mathcal{L}_{b\_aux}$. The total loss is defined as follows:
\begin{equation}
       \mathcal{L}_{total}=\lambda_{basic}\cdot\mathcal{L}_{basic}
        +\lambda_{clip}\cdot\mathcal{L}_{clip}\\
        +\lambda_{boun}\cdot\mathcal{L}_{boun}
        +\lambda_{b\_aux}\cdot\mathcal{L}_{b\_aux},
\end{equation}
where $\mathcal{L}_{basic}$ is basic loss that is commonly used in snippet-level offset regression models \cite{lin2023univtg,video-mamba-suite,cao2024flashvtg} and $\lambda_{basic}$ is weight of basic losses.

The weights of each loss are applied differently to reflect their effects on the training process through experiments. Each loss contributes to learning in different ways and performs different roles, so we adjusted the model to ensure that semantic alignment, boundary prediction, and regression supervision are balanced.
In the inference stage, the saliency score of each clip is calculated based on the clip-specific embeddings generated by the multimodal encoder, and the final timestamp is calculated using the predicted start and end offsets.
\section{Experiments}
\subsection{Setup}
\noindent \textbf{Evaluation metrics}
Performance evaluation follows the official protocol used in the TVR benchmark \cite{xml}. The R1@0.5 and R1@0.7 metrics measure a recall performance of a model, evaluating whether the top 1 prediction segment overlaps with one or more ground truth segments with an IoU of 0.5 or 0.7, respectively. The mAP@0.5 and mAP@0.75 metrics measure the mean average precision at the corresponding IoU thresholds, reflecting both the precision of the predictions and the coverage of the ground truth. mAP@Avg. is the average value of mAP evaluated at intervals of 0.05 from 0.5 to 0.95. Using these metrics, we quantitatively evaluate the accuracy of the model in predicting the answer segment and the boundary prediction accuracy.

\noindent \textbf{Implementation details}
A video encoder, SlowFast\cite{feichtenhofer2019slowfast}, and a text encoder, CLIP\cite{clip}, are used to extract features from video and text, respectively. All experiments are conducted on a single NVIDIA A100 GPU. 
Training is performed using QVHighlights \cite{moment-detr} and TACoS \cite{tacos}, and the weights for the proposed three loss terms are set to 3.0, 3.0, and 2.0, respectively, as shown in Tab.~\ref{tab:ablation_weight}, which demonstrated the best performance. The other hyperparameters follow the settings of the existing baseline models \cite{lin2023univtg,video-mamba-suite,cao2024flashvtg}.

\subsection{Dataset}

\noindent \textbf{QVHighlights}
The QVHighlights dataset \cite{moment-detr} used in this paper is a representative benchmark dataset constructed for moment retrieval that identifies specific time intervals corresponding to natural language queries. The dataset contains 3,478 training videos, 431 validation videos, and 426 test videos, for a total of 4,435 YouTube videos. The total number of query-video pairs is approximately 17,400, and each query has an average of 2.4 answer segments. This means that a single query is able to reference highlight scenes from multiple time points, following a multi-moment annotation structure. In the validation and test sets of the dataset, each instance consists of a 150-second clip extracted from the original video, and the model predicts the answer segment within that video clip.

\noindent \textbf{TACoS}
The TACoS dataset \cite{tacos} is a benchmark for temporal grounding based on the MPII Cooking Composite Activities video corpus. It consists of a total of 127 long cooking activity videos, with 101, 13, and 13 videos allocated to the training, validation, and test sets, respectively. The total number of query-video pairs is 18,835, of which 10,146 are used for training, 4,591 for validation, and 4,098 for testing. The average video length is quite long at approximately 8 minutes, and a single query typically refers to a short segment corresponding to an average of 2.5\% of the total video length. As a result, TACoS is primarily used to evaluate a model's ability to understand detailed and complex activities within long videos and accurately identify temporal segments.

\begin{table}[t]
\caption{\textbf{Moment retrieval results on QVHighlights test split.} All models use the same video encoder (CLIP ViT-B/32 and SlowFast R-50) and text encoder (CLIP text enc.), except for the model marked with $\dagger$, which use InternVideo2 \cite{wang2024internvideo2} as the backbone.}
\vspace{2pt}
\centering
\resizebox{0.6\linewidth}{!}{
\begin{tabular}{l|ccccc}
\toprule
\multirow{2}{*}{\raisebox{-0.8ex}[0pt]{\centering \textbf{Method}}}
 & \multicolumn{2}{c}{\textbf{R1}} & \multicolumn{3}{c}{\textbf{mAP}} \\
\cmidrule(lr){2-3} \cmidrule(lr){4-6}
 & \textbf{@0.5} & \textbf{@0.7} & \textbf{@0.5} & \textbf{@0.75} & \textbf{@Avg.} \\
\midrule
\midrule
CAL \cite{cal} & 25.49 & 11.54 & 23.40 & 7.65 & 9.89 \\
CLIP \cite{clip} & 16.88 & 5.19 & 18.11 & 7.00 & 7.67\\
XML \cite{xml} & 41.83& 30.35& 44.63&31.73 &32.14\\
XML+ \cite{xml} & 46.69& 33.46& 47.89& 34.67 &34.90\\
\midrule
Moment-DETR \cite{moment-detr} & 52.89 &33.02& 54.82 &29.17& 30.73\\
UMT \cite{liu2022umt} &56.23 &41.18& 53.83& 37.01& 36.12\\
QD-DETR \cite{qd-detr} & 62.40 &44.98& 62.52& 39.88 &39.86\\
MomentDiff \cite{momentdiff} &57.42 &39.66& 54.02& 35.73& 35.95\\
TR-DETR \cite{tr-detr} & 64.66 &48.96& 63.98& 43.73 &42.62\\
CG-DETR \cite{cg-detr} & 65.43 &48.38 &64.51 &42.77 &42.86\\
UVCOM \cite{uvcom} &63.55 &47.47& 63.37 &42.67& 43.18\\
\midrule
\midrule
UniVTG \cite{lin2023univtg} & 58.86 &40.86& 57.60& 35.59& 35.47\\
\rowcolor{Gray!20}
+ Ours & 61.16 & 44.32 & 60.37 & 39.94 & 38.08\\
UniVTG w/ PT \cite{lin2023univtg} &65.43& 50.06& 64.06& 45.02& 43.63\\
\rowcolor{Gray!20}
+ Ours & {68.45} & {53.94} & 66.47 & 49.00 & 46.55\\
TD-DETR \cite{td-detr} &64.53& 50.37& 66.21& 47.32& 46.69\\
\rowcolor{Gray!20}
+ Ours & {64.84} & {51.94} & 65.22 & 47.91 & 46.98\\
Video Mamba Suite \cite{video-mamba-suite}& 66.65 &52.19 &64.37& 46.68& 45.18\\
\rowcolor{Gray!20}
+ Ours & 68.00 & 52.65 & 65.65 & 48.11 & 46.23\\
FlashVTG \cite{cao2024flashvtg} & 66.08 & 50.00 & {67.99} & {48.70} & {47.59}\\
\rowcolor{Gray!20}
+ Ours & {68.52} & {53.61} & {68.85} & {52.24} & {50.02}\\
\midrule
FlashVTG$^\dagger$ \cite{cao2024flashvtg} & \underline{70.69} & \underline{53.96} & \underline{72.33} & \underline{53.85} & \underline{52.00}\\
\rowcolor{Gray!20}
+ Ours & \textbf{71.94} & \textbf{57.87} & \textbf{73.03} & \textbf{56.22} & \textbf{53.47}\\
\bottomrule
\end{tabular}
}
\vspace{-15pt}
\label{tab:main_result}
\end{table}
\begin{table}[t]
\centering
\caption{\textbf{Moment retrieval results on TACoS test split. All models use the same video encoder (CLIP ViT-B/32 and SlowFast R-50) and text encoder (CLIP text enc.).}}
\vspace{2pt}
\label{tab:tacos}
\resizebox{0.5\linewidth}{!}{
\begin{tabular}{l|ccc}
\toprule
\textbf{Method} & \textbf{R1@0.3} & \textbf{R1@0.5} & \textbf{R1@0.7} \\
\midrule
\midrule
Moment-DETR \cite{moment-detr} & 37.97 & 24.67 & 11.97 \\
UniVTG \cite{lin2023univtg} & 51.44 & 34.97 & 17.35 \\
CG-DETR \cite{cg-detr} & 52.23 & 39.61 & 22.23 \\
$R^2$-Tuning \cite{r2-tuning} & 49.71 & 38.72 & \underline{25.12} \\
LLMEPET \cite{llmepet} & 52.73 & - & 22.78 \\
\midrule
FlashVTG \cite{cao2024flashvtg} & \textbf{53.71} & \textbf{41.76} & 24.74 \\
\rowcolor{Gray!20}
+ Ours & \underline{53.64} & \underline{41.61} & \textbf{27.49} \\
\bottomrule
\end{tabular}
}
\vspace{-10pt}
\end{table}
\begin{table}[t]
\centering
\caption{\textbf{Effect of loss weights.} Performance change according to different loss weights on QVHighlights.}
\label{tab:ablation_weight}
\vspace{2pt}
\resizebox{0.5\linewidth}{!}{
\begin{tabular}{ccc|ccc}
\toprule
$\lambda_{clip}$ & $\lambda_{bound}$ & $\lambda_{b\_aux}$ & \textbf{R1@0.5} & \textbf{R1@0.7} & \textbf{mAP@Avg.} \\
\midrule
1.0 & 1.0 & 1.0 & 67.35 & 51.48 & 46.79 \\
3.0 & 1.0 & 1.0 & 66.71 & 52.26 & 48.55 \\
1.0 & 3.0 & 1.0 & 67.10 & 51.16 & 48.58 \\
3.0 & 3.0 & 2.0 & \textbf{68.52} & \textbf{53.61} & \textbf{50.02} \\
4.0 & 4.0 & 3.0 & 66.65 & 52.45 & 48.87 \\
4.0 & 5.0 & 4.0 & 67.61 & 52.58 & 48.84 \\
\bottomrule
\end{tabular}
}
\vspace{-10pt}
\end{table}

\subsection{Comparison Results}

\noindent \textbf{QVHighlights}
Tab.~\ref{tab:main_result} shows the results of experiments applying our proposed framework to various moment retrieval models. As shown in Tab.~\ref{tab:main_result}, applying our method to existing models for moment retrieval consistently improves performance. In this paper, we apply our method to four moment retrieval models: UniVTG, TD-DETR, Video Mamba Suite, and FlashVTG. For models using CLIP and SlowFast as backbones, it achieves the highest R1@0.7 when applied to pretrained UniVTG and the highest R1@0.5, mAP@0.5, mAP@0.75 and mAP@Avg. when applied to FlashVTG. In addition, in FlashVTG with the backbone changed to InternVideo2, all metrics achieve higher values than the baseline model. As shown in the experimental results, our proposed method is not limited to specific models and is applicable to various moment retrieval models, demonstrating consistent performance improvements.


\noindent \textbf{TACoS}
Tab.~\ref{tab:tacos} shows the results of experiments applying our proposed framework in TACoS to FlashVTG. As shown in Tab.~\ref{tab:tacos}, our model is effective even with long video datasets such as TACoS. R1@0.3 and R1@0.5 show performance similar to existing baseline models, but R1@0.7, a more accurate metric, shows significantly improved performance. This result indicates that the proposed method is particularly advantageous in capturing precise temporal boundaries.

\subsection{Ablation Studies}

In this section, we analyze the key structural elements that affect the performance of ClipTBP. In our experiments, we utilize FlashVTG \cite{cao2024flashvtg} as a baseline model.
We experimentally verify the effects of weight settings for the three loss terms and the number of top-$k$ selections in the clip-level similarity loss on model performance.

\begin{table}[t]
\centering
\caption{\textbf{Effect of top-$k$ in clip-level alignment loss.} Performance with varying top-$k$ values on QVHighlights.}
\label{tab:ablation_topk}
\vspace{2pt}
\resizebox{0.5\linewidth}{!}{
\begin{tabular}{c|cccc}
\toprule
\textbf{Top-$k$} & \textbf{R1@0.5} & \textbf{R1@0.7} & \textbf{mAP@0.5} & \textbf{mAP@Avg.} \\
\midrule
5 & 67.42 & 52.90 & 67.71 & 48.94 \\
10 & 66.84 & 51.29 & 67.67 & 48.55 \\
20 & \textbf{68.52} & \textbf{53.61} & \textbf{68.85} & \textbf{50.02} \\
30 & 66.77 & 51.16 & 67.83 & 48.68 \\
50 & 66.90 & 52.65 & 67.77 & 48.29 \\
\bottomrule
\end{tabular}
}
\vspace{-18pt}
\end{table}

\begin{table}[t]
\centering
\caption{\textbf{Effect of main boundary loss margin.} Performance change according to different main boundary loss margin on QVHighlights.}
\label{tab:ablation_margin}
\vspace{2pt}
\resizebox{0.6\linewidth}{!}{
\begin{tabular}{ccc|ccc}
\toprule
\makecell{\textbf{Margin} \\ \textbf{threshold}} & 
\makecell{\textbf{Base} \\ \textbf{margin}} & 
\makecell{\textbf{Length} \\ \textbf{scaling}} & 
\textbf{R1@0.5} & 
\textbf{R1@0.7} & 
\textbf{mAP@Avg.} \\
\midrule
0.3 & 0.1 & 0.1 & \textbf{68.52} & \textbf{53.61} & \textbf{50.02} \\
0.2 & 0.1 & 0.1 & 68.52 & 52.71 & 48.58 \\
0.1 & 0.1 & 0.1 & 67.74 & 51.42 & 48.77 \\
0.3 & 0.05 & 0.1 & 64.52 & 48.84 & 46.34 \\
0.3 & 0.2 & 0.1 & 66.71 & 51.35 & 48.37 \\
0.3 & 0.1 & 0.05 & 66.06 & 51.23 & 49.33 \\
0.3 & 0.1 & 0.2 & 68.39 & 53.29 & 49.91 \\
\bottomrule
\end{tabular}
}
\vspace{-18pt}
\end{table}

\noindent \textbf{Effect of loss weights}
Tab.~\ref{tab:ablation_weight} shows the effect of different weight settings on performance when applying the three proposed loss terms: clip-level similarity loss, main boundary loss, and auxiliary boundary loss. In comparison to the base settings where all weights are set to 1.0, increasing $\lambda_{clip}$ to 3.0 improves mAP@Avg., and increasing $\lambda_{bound}$ to 3.0 also improves mAP@Avg. This demonstrates that the two main proposed losses significantly improve moment retrieval performance. In addition to these two losses, increasing $\lambda_{b\_aux}$ to 2.0 achieves the highest mAP@Avg. This result shows that all three losses contribute to improving moment retrieval performance.
When the clip similarity loss weight is too low and query-clip alignment is failed, boundary learning makes unstable predictions based on meaningless segments. In contrast, when boundary loss weight is overly emphasized, semantic alignment becomes inaccurate and the model fails to predict segments at correct positions.

\noindent \textbf{Effect of clip-level alignment top-$k$}
Tab.~\ref{tab:ablation_topk} shows the results of evaluating performance by varying the value of top-$k$ used for hard negative sampling when calculating clip-level similarity loss. When the top-$k$ value is too small, insufficient negative examples are obtained, limiting the learning of semantic relationships. In contrast, when the value is too large, noise increases, leading to a decline in performance. The experimental results show that the best performance is achieved where $k$ is 20, indicating that selecting an appropriate number of hard negatives is effective for improving alignment performance.

\noindent \textbf{Effect of boundary loss margin}
Tab.~\ref{tab:ablation_margin} shows the effects of the three main components of the dynamic margin of the main boundary loss on model performance. The margin threshold determines the upper bound of the dynamically adjusted margin $\delta$, while the base margin defines the minimum margin value applied regardless of segment length, as shown in Eq.~(4). The length scaling term controls how the margin $\delta$ increases as a function of the normalized segment length $(j_e - j_s)/T$, thereby modulating the contribution of segment duration in the boundary loss. 
The results demonstrate that dynamic margin adjustment using length information is effective in improving the performance. Tab.~\ref{tab:ablation_margin} shows that a certain level of dynamic margin adjustment is beneficial, but excessive adjustment causes instability in learning.

\begin{figure*}[!t]
    \centering
    \includegraphics[width=\linewidth]{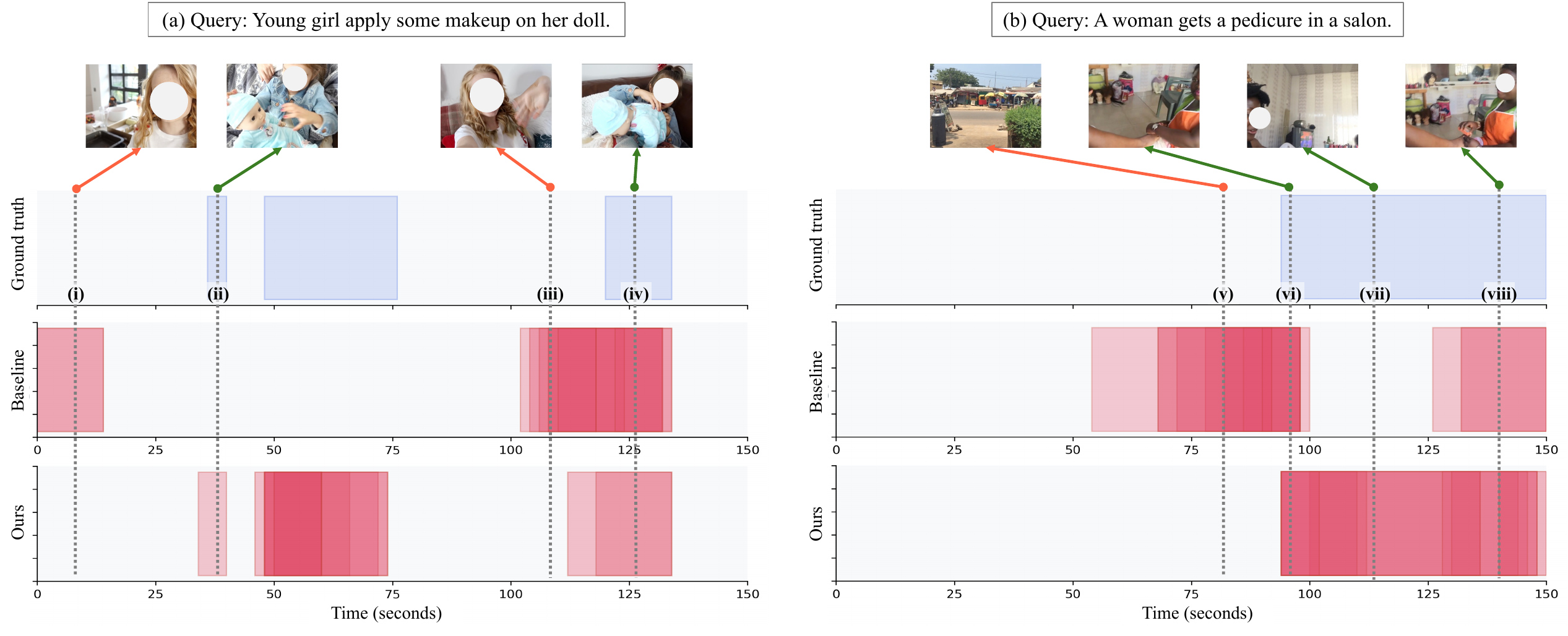}
    \vspace{-20pt}
    \caption{\textbf{Visualization of experimental results for two different example queries using the baseline (FlashVTG \cite{cao2024flashvtg}) and our model on QVHighlights \cite{moment-detr}.} The photos above show part of the connected segment, with the green arrow indicating the correct segment and the red arrow indicating the incorrect segment. The colored sections are the segments predicted by each model. (i), (ii), (iii), (iv), (v): The baseline model is unable to distinguish between semantically unrelated parts in segments with similar visual elements. (vi), (vii), (viii): The three parts belonging to the answer segment are not visually identical, but they all match the query. However, the baseline predicts that (vii), which is visually different from the rest, is not the answer.} 
    \label{visualization_1}
    \vspace{-16pt}
\end{figure*}

\subsection{Qualitative Results}
\noindent \textbf{Visualization of moment retrieval results}
We visualize the results to see how the predictions of our model improve over existing models for actual example text queries for qualitative evaluation. Fig.~\ref{visualization_1} shows the experimental results of our framework combined with FlashVTG using QVHighlights. As shown in Fig.~\ref{visualization_1}(i), (iii), and (v), the baseline model is unable to distinguish between segments that are visually similar to the correct segment but semantically different. As shown in Fig.~\ref{visualization_1}(ii), the baseline model misses the answer segment even for easy segments when there are multiple answer segments. Furthermore, as shown in Fig.~\ref{visualization_1}(vii), the baseline model predicts that some answer segments are not correct when they visually differ from other answer segments, not reflecting their semantic similarity or context. However, our model predicts answer segments that are very similar to GT more accurately than existing models in the various scenarios described above. 
These visualization results demonstrate that our model accurately predicts answers by understanding the context, semantic similarity, and relationships between answer segments. This tendency is consistent with the quantitative evaluations, where Tab.~\ref{tab:main_result} and Tab.~\ref{tab:tacos} show significant improvements in stricter metrics such as R1@0.7, demonstrating that our model achieves more precise boundary prediction than existing methods even in challenging scenarios.

\begin{figure*}[t]
    \centering
    \begin{subfigure}[t]{0.49\linewidth}
        \centering
        \includegraphics[width=\linewidth]{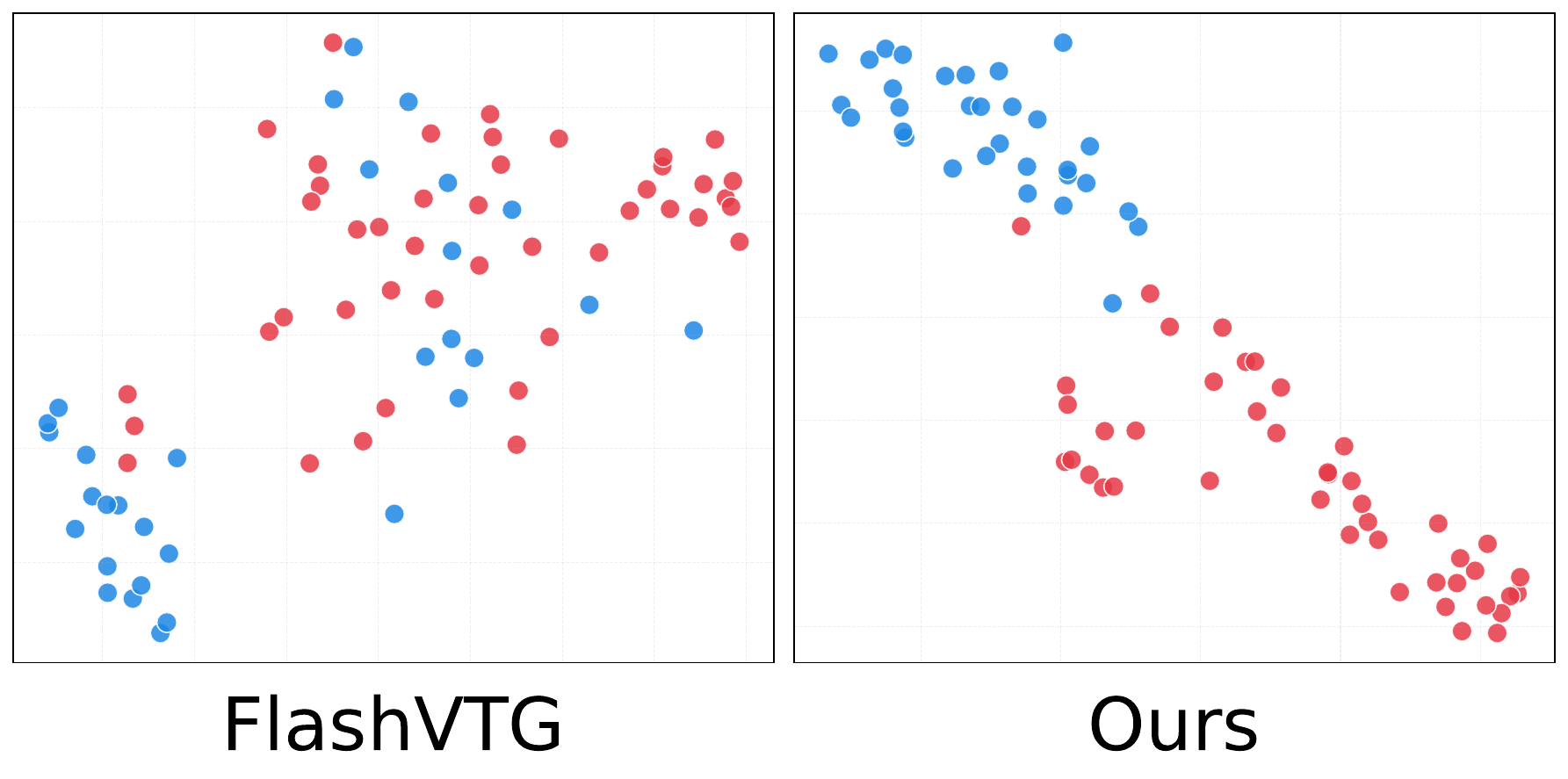}
        \vspace{-20pt}
        \caption{}
    \end{subfigure}
    \hfill
    \begin{subfigure}[t]{0.49\linewidth}
        \centering
        \includegraphics[width=\linewidth]{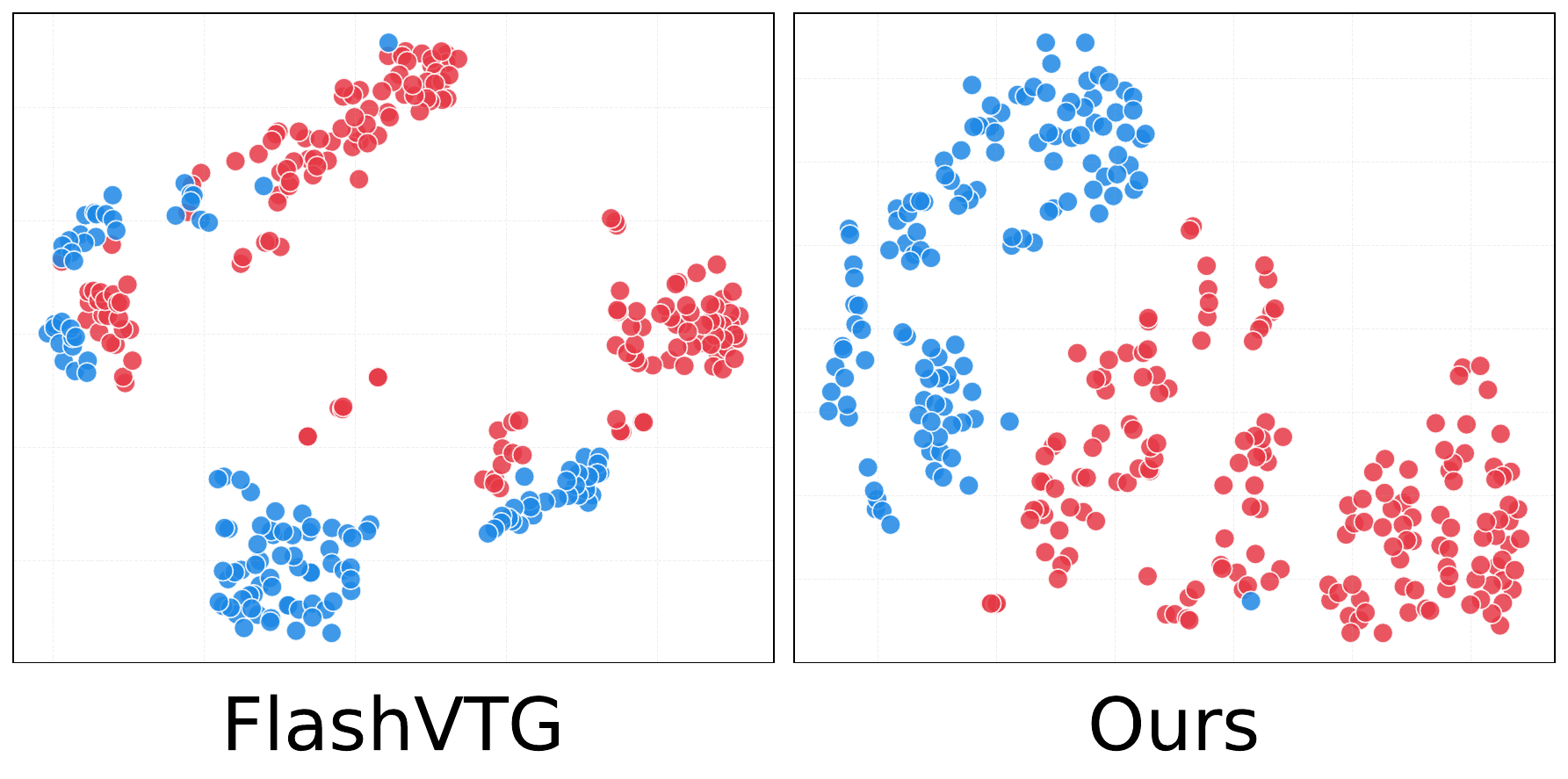}
        \vspace{-20pt}
        \caption{}
    \end{subfigure}
    \vspace{-8pt}
    \caption{\textbf{t-SNE visualization of snippet-level embeddings on QVHighlights.} Blue points denote embeddings of clips that correspond to the ground truth answer segments, while red points denote embeddings of clips that are not part of the ground truth answers and are unrelated to the query. A comparison experiment between FlashVTG \cite{cao2024flashvtg} and our proposed model.}
    \label{tsne fig}
    \vspace{-16pt}
\end{figure*}

\noindent \textbf{t-SNE visualization of snippet-level embeddings} 
Fig.~\ref{tsne fig} visualizes the distributions of snippet-level embeddings generated by FlashVTG and our proposed model for two example videos using t-SNE. As shown in the two results of FlashVTG in Fig.~\ref{tsne fig}, embeddings corresponding to the answer segment and embeddings of segments unrelated to the query are mixed in the embedding space and positioned close to each other. This distribution indicates that segments visually similar to the query or temporally adjacent to the answer segment are mapped to representations similar to those of the answer segment. 
In contrast, embeddings of the answer segments and embeddings of segments excluding the answer are grouped separately in the proposed model, with their boundaries clearly distinguished. This change in distribution demonstrates that our proposed model represents the answer segment and segments unrelated to the query with distinctiveness at the embedding level.

\section{Conclusion}
In this paper, we propose ClipTBP, a clip-pair based temporal boundary prediction framework, to improve moment retrieval performance in videos containing repetitive or ambiguous events. Existing snippet-level offset regression models focus on individual similarities between queries and snippets, which leads to issues such as merging adjacent or similar segments into a single long segment or inaccurately predicting boundaries. ClipTBP addresses these issues by introducing three losses: clip-level similarity loss for learning semantic relationships between answer segments, boundary loss for precise boundary prediction, and auxiliary boundary loss for stabilizing regression. This method improves the accuracy of query-clip alignment and enables more precise prediction of the start and end points of the answer segment. Experiments on several datasets demonstrate consistent performance improvement in various snippet-level models. Ablation results confirm that each loss plays a complementary role and that a balanced combination of clip alignment and boundary regression is essential.

\vspace{5pt}
\noindent \textbf{Acknowledgement}
This work was partly supported by Institute of Information \& Communications Technology Planning \& Evaluation (IITP) under the artificial intelligence graduate school program (Korea University) (No. RS-2019-II190079) and artificial intelligence star fellowship support program to nurture the best talents (IITP-2026-RS-2025-02304828) grant funded by the Korea government (MSIT).

\bibliographystyle{splncs04}
\bibliography{REFERENCE}

@string(NeurIPS = {Adv. Neural Inf. Process. Syst. (NeurIPS)})

@string(CVPR = {Proc. IEEE/CVF Conf. Comput. Vis. Pattern Recognit. (CVPR)})

@string(CVPR_OLD = {Proc. IEEE Conf. Comput. Vis. Pattern Recognit. (CVPR)})

@string(ECCV = {Proc. Eur. Conf. Comput. Vis. (ECCV)})

@string(ICCV = {Proc. IEEE/CVF Int. Conf. Comput. Vis. (ICCV)})

@string(ICML = {Proc. Int. Conf. Machine Learn. (ICML)})

@string(AAAI = {Proc. AAAI Conf. Artif. Intell. (AAAI)})

@string(ICLR = {Int. Conf. Learn. Represent. (ICLR)})

@string(ICASSP = {Proc. IEEE Int. Conf. Acoust. Speech Signal Process. (ICASSP)})

@string(WACV = {Proc. IEEE Winter Conf. Appl. Comput. Vis. (WACV)})

@article{kim2021fre,
  title={Fre-{GAN}: Adversarial frequency-consistent audio synthesis},
  author={Kim, Ji-Hoon and Lee, Sang-Hoon and Lee, Ji-Hyun and Lee, Seong-Whan},
  journal={arXiv preprint arXiv:2106.02297},
  year={2021}
}

@article{jeong2020multimodal,
  title={Multimodal signal dataset for 11 intuitive movement tasks from single upper extremity during multiple recording sessions},
  author={Jeong, Ji-Hoon and Cho, Jeong-Hyun and Shim, Kyung-Hwan and Kwon, Byoung-Hee and Lee, Byeong-Hoo and Lee, Do-Yeun and Lee, Dae-Hyeok and Lee, Seong-Whan},
  journal={GigaScience},
  volume={9},
  number={10},
  pages={giaa098},
  year={2020},
}

@InProceedings{Lee_2020_CVPR,
author = {Lee, Gun-Hee and Lee, Seong-Whan},
title = {Uncertainty-aware mesh decoder for high fidelity 3{D} face reconstruction},
booktitle = CVPR,
year = {2020}
}

@article{MIN2022439,
title = {Attentional feature pyramid network for small object detection},
journal = {Neural Networks},
volume = {155},
pages = {439-450},
year = {2022},
author = {Kyungseo Min and Gun-Hee Lee and Seong-Whan Lee}
}

@inproceedings{wang2024internvideo2,
  title={{InternVideo2}: Scaling foundation models for multimodal video understanding},
  author={Wang, Yi and Li, Kunchang and Li, Xinhao and Yu, Jiashuo and He, Yinan and Chen, Guo and Pei, Baoqi and Zheng, Rongkun and Wang, Zun and Shi, Yansong and others},
  booktitle=ECCV,
  pages={396--416},
  year={2024},
}

@InProceedings{td-detr,
    author    = {Zhou, Xinyang and Wei, Fanyue and Duan, Lixin and Yao, Angela and Li, Wen},
    title     = {The devil is in the spurious correlations: Boosting moment retrieval with dynamic learning},
    booktitle = ICCV,
    year      = {2025},
    pages     = {20981-20990}
}

@inproceedings{luo2024zero,
  title={Zero-shot video moment retrieval from frozen vision-language models},
  author={Luo, Dezhao and Huang, Jiabo and Gong, Shaogang and Jin, Hailin and Liu, Yang},
  booktitle=WACV,
  pages={5464--5473},
  year={2024}
}

@article{cv_1,
  title={Adaptive ambiguity-aware weighting for multi-label recognition with limited annotations},
  author={Shrewsbury, Daniel and Kim, Suneung and Lee, Seong-Whan},
  journal={Neural Networks},
  volume={180},
  pages={106642},
  year={2024},
}

@inproceedings{llmepet,
  title={Prior knowledge integration via llm encoding and pseudo event regulation for video moment retrieval},
  author={Jiang, Yiyang and Zhang, Wengyu and Zhang, Xulu and Wei, Xiao-Yong and Chen, Chang Wen and Li, Qing},
  booktitle={Proc. ACM Int. Conf. Multimedia (ACM MM)},
  pages={7249--7258},
  year={2024}
}

@inproceedings{r2-tuning,
  title={{$R^2$-Tuning}: Efficient image-to-video transfer learning for video temporal grounding},
  author={Liu, Ye and He, Jixuan and Li, Wanhua and Kim, Junsik and Wei, Donglai and Pfister, Hanspeter and Chen, Chang Wen},
  booktitle=ECCV,
  pages={421--438},
  year={2024},
}

@inproceedings{uvcom,
  title={Bridging the {Gap}: A unified video comprehension framework for moment retrieval and highlight detection},
  author={Xiao, Yicheng and Luo, Zhuoyan and Liu, Yong and Ma, Yue and Bian, Hengwei and Ji, Yatai and Yang, Yujiu and Li, Xiu},
  booktitle=CVPR,
  pages={18709--18719},
  year={2024}
}

@article{cg-detr,
  title={Correlation-guided query-dependency calibration for video temporal grounding},
  author={Moon, WonJun and Hyun, Sangeek and Lee, SuBeen and Heo, Jae-Pil},
  journal={arXiv preprint arXiv:2311.08835},
  year={2023}
}

@inproceedings{tr-detr,
  title={{TR-DETR}: Task-reciprocal transformer for joint moment retrieval and highlight detection},
  author={Sun, Hao and Zhou, Mingyao and Chen, Wenjing and Xie, Wei},
  booktitle=AAAI,
  volume={38},
  number={5},
  pages={4998--5007},
  year={2024}
}

@article{momentdiff,
  title={{MomentDiff}: Generative video moment retrieval from random to real},
  author={Li, Pandeng and Xie, Chen-Wei and Xie, Hongtao and Zhao, Liming and Zhang, Lei and Zheng, Yun and Zhao, Deli and Zhang, Yongdong},
  journal=NeurIPS,
  volume={36},
  pages={65948--65966},
  year={2023}
}

@inproceedings{qd-detr,
  title={Query-dependent video representation for moment retrieval and highlight detection},
  author={Moon, WonJun and Hyun, Sangeek and Park, SangUk and Park, Dongchan and Heo, Jae-Pil},
  booktitle=CVPR,
  pages={23023--23033},
  year={2023}
}

@inproceedings{xml,
  title={{TVR}: A large-scale dataset for video-subtitle moment retrieval},
  author={Lei, Jie and Yu, Licheng and Berg, Tamara L and Bansal, Mohit},
  booktitle=ECCV,
  pages={447--463},
  year={2020}
}

@article{cal,
  title={Finding moments in video collections using natural language},
  author={Escorcia, Victor and Soldan, Mattia and Sivic, Josef and Ghanem, Bernard and Russell, Bryan},
  journal={arXiv preprint arXiv:1907.12763},
  year={2019}
}

@inproceedings{ssm,
  title={Efficiently Modeling Long Sequences with Structured State Spaces},
  author={Gu, Albert and Goel, Karan and Re, Christopher},
  booktitle=ICLR,
  year={2022}
}

@inproceedings{vim,
author = {Zhu, Lianghui and Liao, Bencheng and Zhang, Qian and Wang, Xinlong and Liu, Wenyu and Wang, Xinggang},
title = {{Vision Mamba: Efficient} visual representation learning with bidirectional state space model},
year = {2024},
booktitle = ICML,
articleno = {2584},
numpages = {14}
}

@inproceedings{panta2024cross,
  title={Cross-modal contrastive learning with asymmetric co-attention network for video moment retrieval},
  author={Panta, Love and Shrestha, Prashant and Sapkota, Brabeem and Bhattarai, Amrita and Manandhar, Suresh and Sah, Anand Kumar},
  booktitle=WACV,
  pages={607--614},
  year={2024}
}

@inproceedings{detr,
  title={End-to-end object detection with transformers},
  author={Carion, Nicolas and Massa, Francisco and Synnaeve, Gabriel and Usunier, Nicolas and Kirillov, Alexander and Zagoruyko, Sergey},
  booktitle=ECCV,
  pages={213--229},
  year={2020},
}

@inproceedings{moment-detr,
  title={Detecting moments and highlights in videos via natural language queries},
  author={Lei, Jie and Berg, Tamara L and Bansal, Mohit},
  booktitle=NeurIPS,
  volume={34},
  pages={11846--11858},
  year={2021}
}

@inproceedings{liu2022umt,
  title={{UMT}: Unified multi-modal transformers for joint video moment retrieval and highlight detection},
  author={Liu, Ye and Li, Siyuan and Wu, Yang and Chen, Chang-Wen and Shan, Ying and Qie, Xiaohu},
  booktitle=CVPR,
  pages={3042--3051},
  year={2022}
}

@article{cao2024flashvtg,
  title={{FlashVTG}: Feature layering and adaptive score handling network for video temporal grounding},
  author={Cao, Zhuo and Zhang, Bingqing and Du, Heming and Yu, Xin and Li, Xue and Wang, Sen},
  journal={arXiv preprint arXiv:2412.13441},
  year={2024}
}

@article{video-mamba-suite,
  title={Video mamba suite: State space model as a versatile alternative for video understanding},
  author={Chen, Guo and Huang, Yifei and Xu, Jilan and Pei, Baoqi and Chen, Zhe and Li, Zhiqi and Wang, Jiahao and Li, Kunchang and Lu, Tong and Wang, Limin},
  journal={arXiv preprint arXiv:2403.09626},
  year={2024}
}

@inproceedings{lin2023univtg,
  title={{UniVTG}: Towards unified video-language temporal grounding},
  author={Lin, Kevin Qinghong and Zhang, Pengchuan and Chen, Joya and Pramanick, Shraman and Gao, Difei and Wang, Alex Jinpeng and Yan, Rui and Shou, Mike Zheng},
  booktitle=ICCV,
  pages={2794--2804},
  year={2023}
}

@inproceedings{feichtenhofer2019slowfast,
  title={Slowfast networks for video recognition},
  author={Feichtenhofer, Christoph and Fan, Haoqi and Malik, Jitendra and He, Kaiming},
  booktitle=ICCV,
  pages={6202--6211},
  year={2019}
}

@InProceedings{clip,
  title = 	 {Learning transferable visual models from natural language supervision},
  author =       {Radford, Alec and Kim, Jong Wook and Hallacy, Chris and Ramesh, Aditya and Goh, Gabriel and Agarwal, Sandhini and Sastry, Girish and Askell, Amanda and Mishkin, Pamela and Clark, Jack and Krueger, Gretchen and Sutskever, Ilya},
  booktitle = ICML,
  pages = 	 {8748--8763},
  year = 	 {2021},
  volume = 	 {139}
}

@inproceedings{mr_1,
  title={Learning 2d temporal adjacent networks for moment localization with natural language},
  author={Zhang, Songyang and Peng, Houwen and Fu, Jianlong and Luo, Jiebo},
  booktitle=AAAI,
  volume={34},
  number={07},
  pages={12870--12877},
  year={2020}
}

@inproceedings{mr_2,
  title={Local-global video-text interactions for temporal grounding},
  author={Mun, Jonghwan and Cho, Minsu and Han, Bohyung},
  booktitle=CVPR,
  pages={10810--10819},
  year={2020}
}

@inproceedings{mr_3,
  title={Dual attention networks for multimodal reasoning and matching},
  author={Nam, Hyeonseob and Ha, Jung-Woo and Kim, Jeonghee},
  booktitle=CVPR_OLD,
  pages={299--307},
  year={2017}
}

@article{mr_4,
  title={Text-to-clip video retrieval with early fusion and re-captioning},
  author={Xu, Huijuan and He, Kun and Sigal, Leonid and Sclaroff, Stan and Saenko, Kate},
  journal={arXiv preprint arXiv:1804.05113},
  volume={2},
  number={6},
  pages={7},
  year={2018}
}

@inproceedings{contrastive_learning,
  title={Dimensionality reduction by learning an invariant mapping},
  author={Hadsell, Raia and Chopra, Sumit and LeCun, Yann},
  booktitle=CVPR_OLD,
  volume={2},
  pages={1735--1742},
  year={2006}
}

@inproceedings{triplet_loss,
  title={{FaceNet}: A unified embedding for face recognition and clustering},
  author={Schroff, Florian and Kalenichenko, Dmitry and Philbin, James},
  booktitle=CVPR_OLD,
  pages={815--823},
  year={2015}
}

@article{tacos,
  title={Grounding action descriptions in videos},
  author={Regneri, Michaela and Rohrbach, Marcus and Wetzel, Dominikus and Thater, Stefan and Schiele, Bernt and Pinkal, Manfred},
  journal={Trans. of the Assoc. for Comput. Linguistics (TACL)},
  volume={1},
  pages={25--36},
  year={2013}
}

@inproceedings{smooth_l1,
  title={Fast {R-CNN}},
  author={Girshick, Ross},
  booktitle=ICCV,
  pages={1440--1448},
  year={2015}
}

@inproceedings{pan2025fawl,
  title={{FAWL}: Weakly-Supervised Video Corpus Moment Retrieval with Frame-Wise Auxiliary Alignment and Weighted Contrastive Learning},
  author={Pan, Yi and Zhang, Yujia and Zhao, Xiaoguang},
  booktitle=ICASSP,
  pages={1--5},
  year={2025},
}

@article{zhang2021video,
  title={Video Corpus Moment Retrieval with Contrastive Learning},
  author={Zhang, Hao and Sun, Aixin and Jing, Wei and Nan, Guoshun and Zhen, Liangli and Zhou, Joey Tianyi and Goh, Rick Siow Mong},
  journal={arXiv preprint arXiv:2105.06247},
  year={2021}
}

\end{document}